\documentclass[sigconf,table,xcdraw]{acmart}

\usepackage[utf8]{inputenc}
\usepackage{natbib}
\setcitestyle{authoryear,open={(},close={)}} 
\usepackage{amsmath,amsfonts}
\usepackage{graphicx}
\usepackage{textcomp}
\def\BibTeX{{\rm B\kern-.05em{\sc i\kern-.025em b}\kern-.08em
    T\kern-.1667em\lower.7ex\hbox{E}\kern-.125emX}}

\graphicspath{ {./figures/} }
\usepackage{algorithm}
\usepackage{algpseudocode}
\usepackage{algpseudocode}
\usepackage{subcaption}
\usepackage{multirow}
\usepackage{hyperref}
\usepackage{graphicx}
\usepackage{adjustbox}
\usepackage[justification=centering]{caption}

\AtBeginDocument{%
  \providecommand\BibTeX{{%
    \normalfont B\kern-0.5em{\scshape i\kern-0.25em b}\kern-0.8em\TeX}}}

\setcopyright{acmcopyright}
\copyrightyear{2022}
\acmYear{2022}

\acmConference[MiLeTS '22, SIGKDD]{MiLeTS '22}{August 15, 2022}{Washington DC, USA}

\begin{document}

\title{Shapelet-Based Counterfactual Explanations for Multivariate Time Series}

\author{Omar Bahri}
\email{omar.bahri@usu.edu}
\affiliation{%
  \institution{Utah State University}
  \city{Logan}
  \state{Utah}
  \country{USA}
}
\author{Soukaina Filali Boubrahimi}
\email{soukaina.boubrahimi@usu.edu}
\affiliation{%
  \institution{Utah State University}
  \city{Logan}
  \state{Utah}
  \country{USA}
}
\author{Shah Muhammad Hamdi}
\email{shamdi1@nmsu.edu}
\affiliation{%
  \institution{New Mexico State University}
  \city{Las Cruces}
  \state{New Mexico}
  \country{USA}
}

\renewcommand{\shortauthors}{Bahri et al.}

\begin{abstract}
As machine learning and deep learning models have become highly prevalent in a multitude of domains, the main reservation in their adoption for decision-making processes is their black-box nature. The Explainable Artificial Intelligence (XAI) paradigm has gained a lot of momentum lately due to its ability to reduce models opacity. XAI methods have not only increased stakeholders' trust in the decision process but also helped developers ensure its fairness. Recent efforts have been invested in creating transparent models and post-hoc explanations. However, fewer methods have been developed for time series data, and even less when it comes to multivariate datasets. In this work, we take advantage of the inherent interpretability of shapelets to develop a model agnostic multivariate time series (MTS) counterfactual explanation algorithm. Counterfactuals can have a tremendous impact on making black-box models explainable by indicating what changes have to be performed on the input to change the final decision. We test our approach on a real-life solar flare prediction dataset and prove that our approach produces high-quality counterfactuals. Moreover, a comparison to the only MTS counterfactual generation algorithm shows that, in addition to being visually interpretable, our explanations are superior in terms of proximity, sparsity, and plausibility.
\end{abstract}



\maketitle

\section{Introduction}
During the last decade, machine learning and deep learning models have established themselves as the state-of-the-art in multiple scientific and data-centric domains. Their high performance in prediction and classification tasks has allowed them to be adopted in a plethora of fields such as banking, insurance, healthcare, and meteorology \cite{Angra2017,Shinde2018,Sarker2021}. Despite this huge success, one of the main issues still hindering their full deployment is the black-box nature of most algorithms, and the limitations it imposes in terms of interpretability and explainability. This in turn brings on questions about the fairness of such algorithms, making them difficult to include in critical decision-making processes. In this context, the EU General Data Protection Regulation \cite{GDPR}, a law that insists on the importance of fairness, trustworthiness, and privacy, and that urges companies to provide explanations to users and consumers, has been introduced in 2016. In addition, initiatives such as DARPA's Explainable AI (XAI) \cite{Gunning2016} have been started by different agencies to develop interpretable machine learning solutions. As a result, more transparent machine learning algorithms such as decision trees and linear regression have seen increased interest. In parallel, a novel line of research that focuses on providing post hoc interpretable explanations generated by extra modules on top of the complex black-box models was met with great success. While most of these methods focused on image data \cite{VanLooveren2019}, tabular data \cite{VanLooveren2019,Lundberg2017,Schlegel2019}, and text data \cite{Ribeiro2016a}, more recent works started tackling univariate time series \cite{Guidotti2020,Parvatharaju2021a,Delaney2020,Ates}. However, because of their high-dimensionality, multivariate time series (MTS) data remain under-explored, with CoMTE \cite{Ates} being the only counterfactual explanation method specifically designed for MTS. In this work, we capitalize on the inherent interpretability of shapelet-based algorithms in time series classification (TSC) to develop a shapelet explainer for time series (SETS), a model-agnostic counterfactual generation algorithm for MTS data. We evaluate our results on an open-source solar-flare dataset. The rest of this paper is organized as follows. In Section 2, we present the state-of-the-art machine learning explanation methods, with a particular focus on time series. In Section 3, we introduce SETS and discuss some important counterfactual evaluation measures. In Section 4, we present the experimental setup. In Section 5, we discuss the results. And in Section 6, we conclude with a summary.

\section{Related Work}
One of the most popular model-agnostic explanation methods is LIME, developed by \citet{Ribeiro2016a}. To explain a dataset sample, LIME generates random neighboring instances around it by performing small perturbations. Then, a surrogate linear model that mimics the behavior of the original black-box model is trained on the generated instances, and its feature importance values are used to explain the model decision. One of the main drawbacks of LIME is its assumption of linearity, which rarely holds true when it comes to high-dimensional and complex time series datasets \cite{Parvatharaju2021a}. Inspired by game theory, SHAP \cite{Lundberg2017} is another feature-based explanation method that overcomes this linearity limitation. It computes the importance of features by deriving their additive Shapley values. Although more robust than LIME, both methods raise concerns about their stability, as slight perturbations can totally change the model decision \cite{Adebayo2018,Nguyen2020a,Delaney2020}. Instance-based explanation methods were developed as an alternative to feature-based approaches. In particular, counterfactuals, artificial instances generated as close as possible to the original dataset sample in such a way that the model prediction changes have gained increased popularity. In general, counterfactuals are generated by introducing perturbations from a representative sample of the target class or guided by an objective function (or both). \citet{Wachter2018} were among the first to generate counterfactual explanations. Their method consists in minimizing a loss function consisting of a prediction term to reach the target class label, and a distance term to ensure the counterfactual lies close to the original instance. \citet{Mothilal2020} extend this approach by adding a diversity constraint to allow the generation of different counterfactuals. \citet{Dhurandhar2018} introduced an autoencoder-based loss term to enforce interpretability by keeping the counterfactuals within the target class data distribution. 
In addition, \citet{VanLooveren2019} add a prototype loss term to ensure interpretability and speed up the search process. Even though none of the techniques mentioned so far have been proposed for time series data, they might in theory be used in this context. For example, an apriori segmentation of the data --although it results in some information loss-- has made it possible for SHAP and LIME to be applied to time series \cite{Schlegel2019}. In the work by \citet{Ates}, 11 statistical features have been extracted from the data. However, the results have not been satisfactory \cite{Ates}. Similarly, instance-based explanation methods do not yield the best results when employed for time series \cite{Delaney2020}. More recently, a few techniques designed for time series datasets have made their way to the literature. \citet{Guidotti2020} build a decision tree using shapelets \cite{Grabocka2014} extracted from the dataset. Then, they extract shapelet-based rules from the decision tree to explain the black-box model decisions. In native guide (NG) \cite{Delaney2020}, the original instance's nearest-unlike-neighbor (nun) is extracted from the target class and used to perturb the original instance. When possible, the most important contiguous subsequence is found using Class Activation Mapping \cite{Zhou2015}, and the perturbations are introduced at its level. Otherwise, dynamic time warping barycenter averaging (DBA) \cite{Petitjean2011} is used. \citet{Ates} introduced CoMTE, a counterfactual multivariate time series explainability method. CoMTE extracts a nun for each feature variable using KD-trees, and replaces entire dimensions to build a counterfactual. The authors proposed a heuristic search method based on hill climbing to select the feature variables to be modified. In case the heuristic fails to provide a counterfactual, a greedy search is performed to find the optimal feature set. To our knowledge, this is the first counterfactual generation algorithm developed for MTS.

\section{Methodology}

\subsection{Shapelet Transform}
A shapelet is a phase-independent, characteristic subsequence that occurs repeatedly in a time series dataset. Since the first shapelet-based classification algorithm was introduced by \citet{Ye2011}, multiple works followed suit \cite{Rakthanmanon2013,Grabocka2014,Lines2012,Fang2018}. In a recent time series classification benchmark study by \citet{Bagnall2017}, shapelet transform (ST) \cite{Lines2012,Hills2014,Bostrom2015a} proved to be among the best algorithms. Therefore, we use it to extract shapelets for our algorithm. Moreover, shapelet discovery and classification in ST are performed in two separate steps, which makes its choice even more convenient.\\
The first step involved in mining shapelets using ST is to extract all candidate shapelets  $S_i$ from each dataset sample, i.e. all subsequences of the desired predefined lengths. In the next step, each $S_i$ is slided across each dataset sample, and the minimum distance that separates it from all subsequences $w$ of similar length is recorded as the distance to that dataset sample. Equation 1 defines the sliding window function, such as the set of all subsequences of length equal to that of $S$ is represented by $W$. Then, the final set of shapelets is selected based on their information gain.

\begin{equation}\label{sliding_window}
sDist(S,T)=min_{w\in W}(dist(S,w))
\end{equation}

\subsection{Shapelet Explainer for Time Series}
In this section, we present SETS, an instance-based, model-agnostic counterfactual generation algorithm.

\subsubsection{Model Fitting}
Given the inherent interpretability of shapelets and the success of shapelet-based algorithms, we consider the shapelets extracted using ST as the building blocks of our counterfactual generation algorithm. During the computation of the distances separating shapelets from dataset instances, we store the distances between every shapelet and its potential occurrences (subsequences of the same length) and retain the closest ones according to a predefined threshold. These retained shapelets occurrences are then used to select class-shapelets: shapelets characteristic of each dataset class, i.e. those that happen under that class only. The remaining shapelets are discarded. Then, for each shapelet, the occurrence distribution is computed as the average of all its occurrences.\\

\subsubsection{Counterfactual Generation}

\begin{figure*}[t!]
    \centering
    \includegraphics[width=\textwidth,height=\textheight,keepaspectratio]{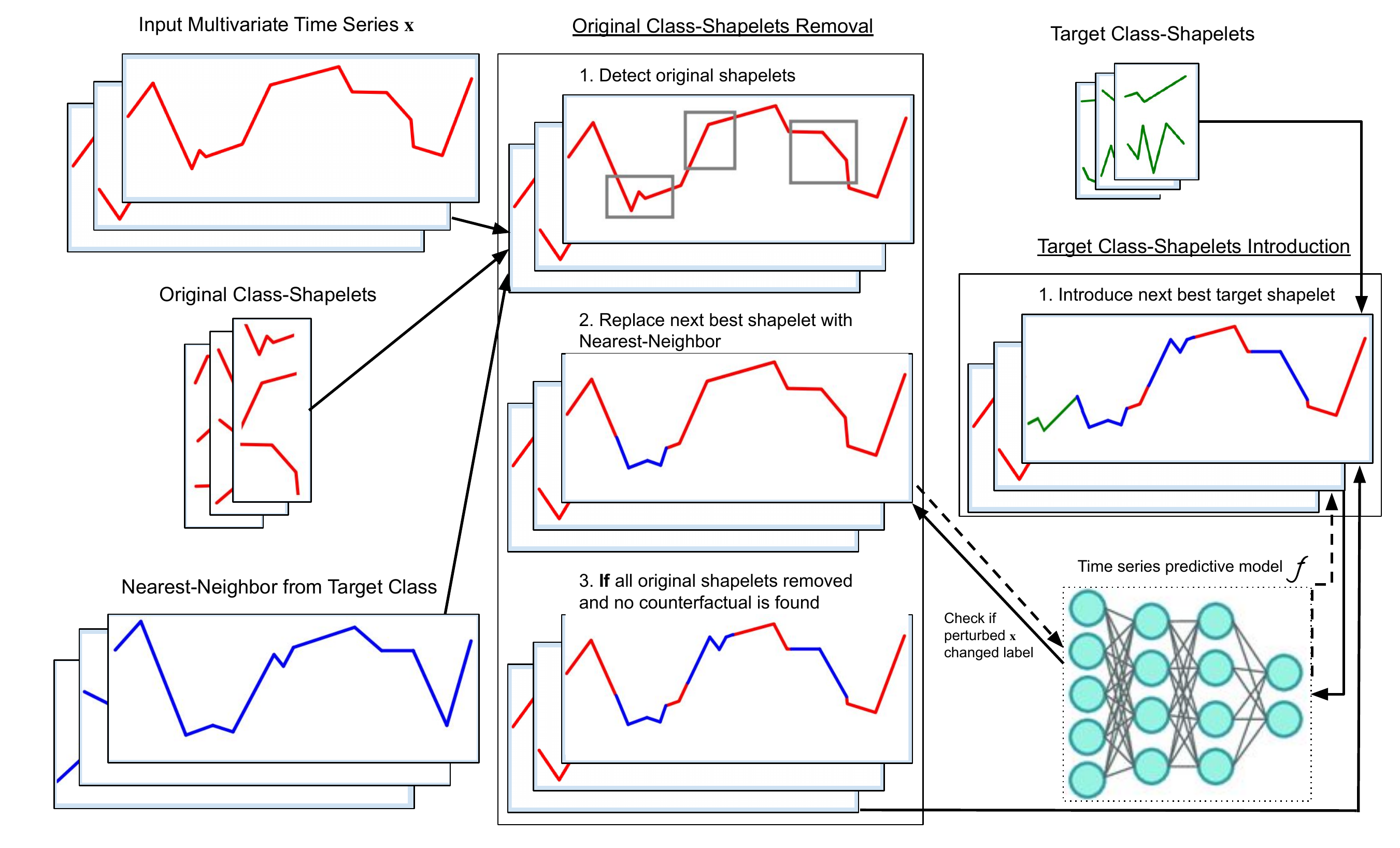}
  \caption{SETS single dimension perturbation process}
\end{figure*}

Given a dataset instance $x$ of class $A$, generating a counterfactual instance $x_{cf}$ of class $B$ starts with the extraction of the nearest-neighbor $x_{nn}$ of $x$ from class $B$ instances using the k-Nearest-Neighbor (kNN) algorithm. Next, the dataset dimensions are sorted in descending order according to their highest scoring shapelets. Then, steps $a$ and $b$ below are performed for each dimension, class-shapelet by class-shapelet and until a valid counterfactual $x_{cf}$ is found, i.e. until $argmax(f(x_{cf}))=B$, where $f$ is the prediction function of the black-box model. Figure 1. illustrates the process for a single dimension. If none of the dimensions succeed in creating a valid counterfactual, the perturbations from all possible subsets of dimensions are combined until a valid counterfactual is found.

\paragraph{Original Class-Shapelets Removal}
$x_{cf}$ is constructed by replacing each $A$ class-shapelet --by descending order of information gain-- contained in $x$ by the values of $x_{nn}$ at the same time steps, scaled to the original feature range using min-max scaling. The motivation behind this step is to remove the shapelets that swayed the model prediction $f(x_{cf})$ toward label $A$.

\paragraph{Target Class-Shapelets Introduction}
$x_{cf}$ is constructed by introducing each $B$ class-shapelet --by descending order of information gain-- into the time steps defined by its occurrence distribution, scaled to the original feature range using min-max scaling. The motivation behind this step is to influence the model prediction $f(x_{cf})$ towards label $B$.

\subsection{Evaluation Measures}
By relying on shapelets in the counterfactual generation process, we are able to build highly interpretable explanations. On the one hand, performing changes at the level of shapelets only creates meaningful perturbations. This adds value to the counterfactuals by making them easily interpretable by stakeholders. On the other hand, being able to visualize the shapelets and see their impact on tasks such as classification and prediction, independently of the counterfactuals, increases the confidence in the SETS algorithm. However, there is still a need for quantitative evaluation, particularly when it comes to comparing different counterfactual generation approaches. Although there is no standard to evaluate explanation methods \cite{Ates,Schmidt2019,Lipton2016}, proximity, interpretability, and sparsity have been used repeatedly in the literature \cite{VanLooveren2019,Karimi2019,Delaney2020,Ates,Mothilal2020}. In this section, we introduce the three measures and describe how we compute them to evaluate our approach.

\subsubsection{Proximity}
Also referred to as distance or closeness, the proximity measure ensures that the counterfactual is close to the original sample. Ideally, the perturbations should be as small as possible. However, proximity is not the only criteria sought in a good counterfactual. Therefore, a balance has to be found with the two measures below. Following \cite{Downs2020,Karimi2019,Keane2021,Delaney2020}, we decided to use three distance metrics to evaluate the proximity of counterfactuals. We use the Manhattan distance ($L_{1}$-norm, equation 2) and the Euclidian distance ($L_{2}$-norm, equation 3) to measure the distance between the counterfactual $x_{cf}$ and its original instance $x$, and the $L_{\inf}$-norm (equation 3) to get the magnitude of the highest perturbation at a single time step. $D$ is the number of dimensions in the multivariate dataset, and $T$ is the length of the series.

\begin{equation}\label{l1}
\left|\left|x-x_{cf}\right|\right|_{L_1} = \sum_{i}^D \sum_{j}^T  \left|x_{ij}-{x_{cf}}_{ij}\right|
\end{equation}

\begin{equation}\label{l2}
\left|\left|x-x_{cf}\right|\right|_{L_2} = \sqrt{ \sum_{i}^D \sum_{j}^T  \left(x_{ij}-{x_{cf}}_{ij}\right)^2 }
\end{equation}

\begin{equation}\label{linf}
\left|\left|x-x_{cf}\right|\right|_{L_{inf}} = \sum_{i}^D \sum_{j}^T max\left|x_{ij}-{x_{cf}}_{ij}\right|
\end{equation}

\subsubsection{Sparsity}
Another important quality in a good counterfactual is the sparsity of the perturbed features and in the case of time series, of the perturbed time steps. Introducing changes to several features and at different time steps will not only make the explanation harder to comprehend by stakeholders but might even make it impossible to perform. Therefore, perturbed features should be kept to a minimum. In addition, time series data should be changed at the level of short, contiguous intervals, as perturbations that affect single, dispersed time steps are not meaningful \cite{Parvatharaju2021a,Delaney2020}.

\subsubsection{Plausibility}
Being close to the original sample and having sparse perturbations might not be enough to warrant the plausibility of a counterfactual. Counterfactual explanations need to be easily interpretable by humans. Thus, they must be realistic. One way to quantitatively study the plausibility of a counterfactual is to check whether it belongs to the same data manifold of the dataset \cite{VanLooveren2019,Delaney2020,Poyiadzi2019,Karimi2019}. This can be achieved by applying novelty detection techniques, which detect out-of-distribution (OOD) instances.\\
Similarly to the work by \citet{Delaney2020}, we adopt three novelty detection approaches to assess the plausibility of counterfactuals. In particular, we perform (1) the local outlier factor (LOF) method \cite{Breuniq2000,Kanamori2020} which calculates the local density deviation of each instance with respect to its neighbors and flags the ones with lower densities as outliers, (2) isolation forest (IF) \cite{Liu2008} which considers how far a dataset instance is to the rest of the dataset, and the one class support vector machine (OC-SVM) \cite{Scholkopf2001} method (on the raw time and the matrix profile (OC-SVM MP) \cite{Yeh2017} representations of the time series).

\section{Experimental Setup}

\subsection{Dataset}
A solar flare is an extremely powerful burst of electromagnetic radiation originating from the surface of the sun. Because of their sporadicity, forecasting them remains a big challenge. Furthermore, solar flares are highly dangerous for astronauts, space equipment, and can even damage infrastracture at ground surface such as electric power grids and navigational signals \cite{Boubrahimi2018a,Martens2017}. Therefore, forecasting solar flare events is very important to perform critical preventive measures and save billions of dollars worth of damage \cite{Angryk2020}. Since solar flares are rare events, the datasets available for training forecasting models are small, which makes their outputs less dependable. Thus, generating post hoc explanations in the form of counterfactuals has the potential of increasing their trustworthiness in the eyes of stakeholders, including physicists and decision makers.\\
We evaluate SETS on a real-life, MTS solar flare dataset created by a research group from Georgia State University \cite{Angryk2020}. The dataset contains 1354 samples of 60 time steps, recorded at 12 minutes intervals for a total of 12 hours. The class distribution of the dataset samples is shown in Table 1. The four class labels represent the classification of the most powerful solar flare recorded in the following 12 hours. The data was captured by the Helioseismic Magnetic Imager (HMI) \cite{Schou2012,Bobra2014,Angryk2020a} on the Solar Dynamics Observatory (SDO) \cite{Pesnell2012} run by NASA.

\begin{table}\label{dataset}
\centering
\caption{Dataset Class Distribution}
\begin{tabular}{ |c|c|c|c|c|c| }
 \hline
 \bf{Class} & \bf{X} & \bf{M} & \bf{B/C} & \bf{Q} & \bf{All} \\ 
 \hline
 \bf{Number of elements} & 303 & 350 & 356 & 345 & 1354 \\ 
 \hline
\end{tabular}
\end{table}

\subsection{Implementation Details}
We used the sktime \cite{Loning2019} implementation of ST, and introduced a slight modification to extract the indices of the occurrences of each shapelet along with their distances as described in Section 3.2. Because of the high-dimensionality of the dataset, we ran the contracted shapelet transform implementation, which does not significantly hurt the performance of ST \cite{Bostrom2017}. This approach consists in randomly selecting shapelets for a user-defined amount of time, instead of trying all possible subsequences. For this experiment, we ran the algorithm for a total of 4 hours, including all dimensions. In order to achieve sparse perturbations, the length of the shapelets was restricted to a maximum of 50\% of the length of the time series. We provide access to our code and to the solar-flare dataset on our \href{https://github.com/omarbahri/SETS}{{\color{blue} GitHub repository}}.\\

\subsection{Compared Methods}
\subsubsection{NG}
It extracts the nearest neighbor to the original instance from the target class and introduces perturbations from its time steps. We employ the model agnostic version of the algorithm, which uses DBA to guide the perturbations. The implementation is provided by the authors in the original publication \citet{Delaney2020}. Since NG was developed for univariate time series datasets, we adapt it to the multivariate case by simply reshaping the entire dataset under one dimension.

\subsubsection{CoMTE}
To our knowledge, CoMTE \cite{Ates} is the only counterfactual explanation method developed specifically for MTS. First, CoMTE finds a distractor from the target class by constructing its KD-tree and considering the original instance's nearest neighbor. Then, it picks a small set of feature dimensions from the original sample and replaces it with the distractor's variables, resulting in a counterfactual with the target class label. Ideally, the set of feature dimensions should be optimal, which can be achieved using a greedy search. To speed up the search process, the authors proposed a heuristic method based on hill climbing, followed by a post hoc trimming step. In case the heuristic fails to provide a counterfactual, the greedy search is performed. The implementation we use is provided in the original publication.

\section{Experimental Results}
We split the solar flare dataset into a training set with 70\% of the instances and a testing set with the remaining 30\%. Then, we run ST on the training set as described in section 4.2, and use the extracted shapelets to create our explanations as detailed in section 3.2. Since the dataset contains 4 classes, we generate 3 counterfactuals for each instance --one for each of the possible target classes-- for a total of 1221 counterfactuals. While we use a simple neural network as the black-box model, NG, CoMTE, and SETS are model agnostic and can be applied to any other machine learning model.

\subsection{Qualitative Evaluation}
SETS starts by finding class-shapelets, i.e. shapelets that occur under one class only. By first examining these shapelets, the user can have a better understanding of their discriminative power and their role in the dataset. Domain experts can even go a step further and draw important insights into the problem at hand. Figure 2 shows two shapelets that happen uniquely under class X, at the level of the total unsigned current helicity feature. This means that whenever one of these two shapelets is detected, an extremely powerful solar-flare event is going to burst.

\begin{figure}
    \centering
    \begin{subfigure}[t]{0.5\columnwidth}
        \centering
        \includegraphics[height=3.3cm]{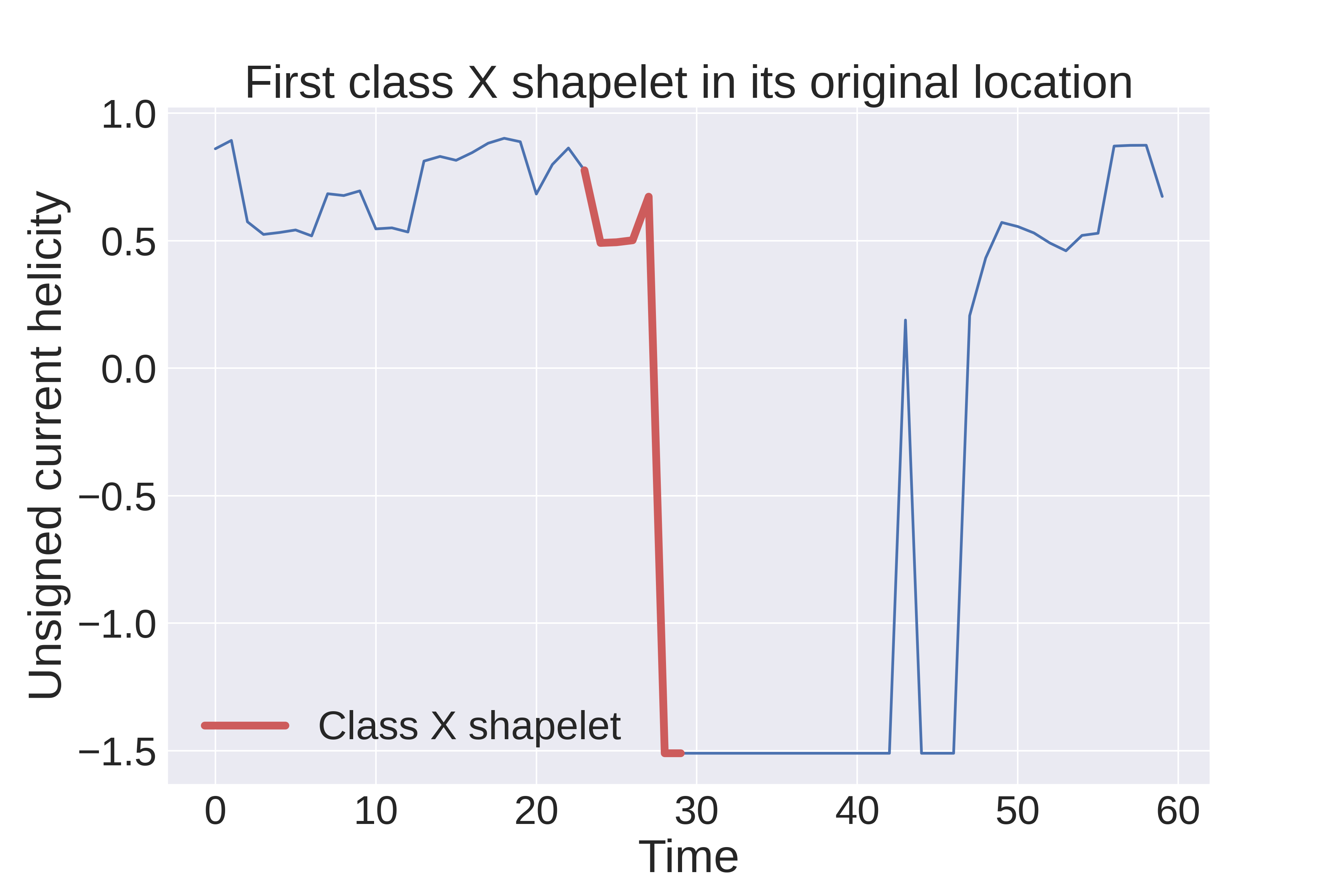}
        \caption{}
    \end{subfigure}%
    ~ 
    \begin{subfigure}[t]{0.5\columnwidth}
        \centering
        \includegraphics[height=3.3cm]{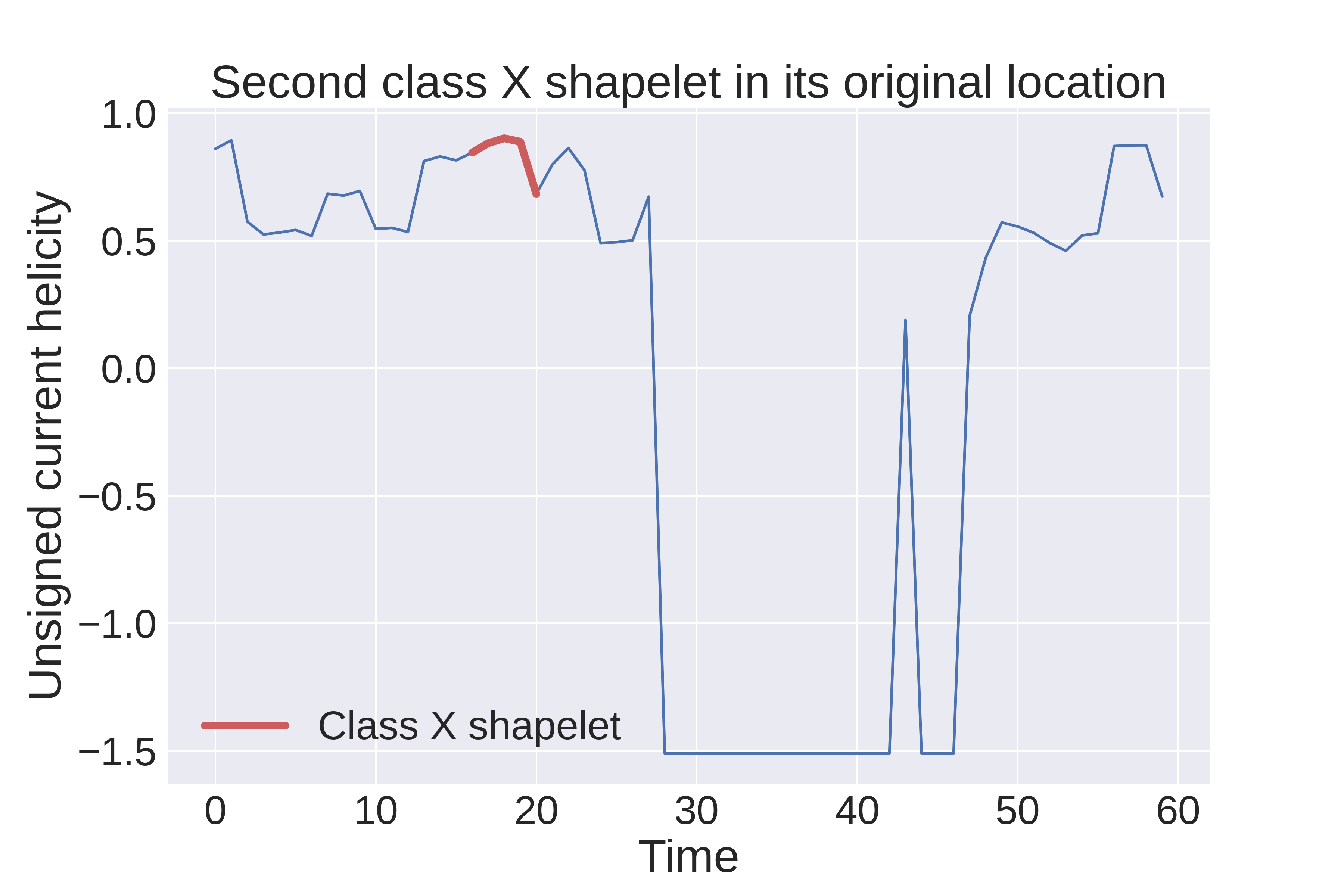}
        \caption{}
    \end{subfigure}
    \caption{Class X Shapelets}
\end{figure}

Then, SETS exploits the class-shapelets to create meaningful counterfactual explanations. For example, considering the unsigned current helicity dimension of the original time series sample in Figure 3.a, SETS generates a counterfactual of class X (from the original class M) by simply introducing the shapelet from Figure 2.a. The perturbed dimension of the counterfactual is plotted in Figure 3.b.

\begin{figure}
    \centering
    \begin{subfigure}[t]{0.5\columnwidth}
        \centering
        \includegraphics[height=3.2cm]{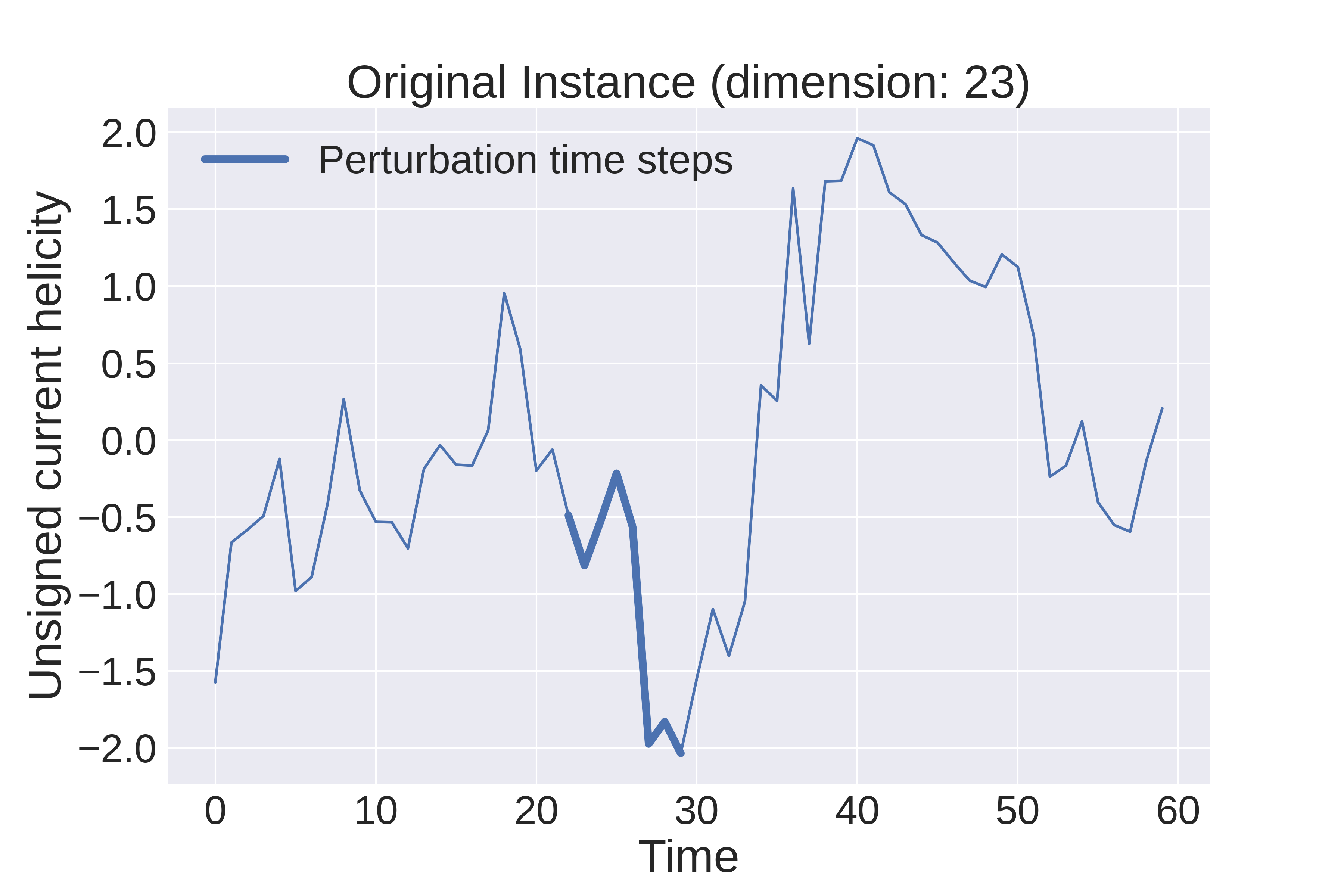}
        \caption{}
    \end{subfigure}%
    ~ 
    \begin{subfigure}[t]{0.5\columnwidth}
        \centering
        \includegraphics[height=3.2cm]{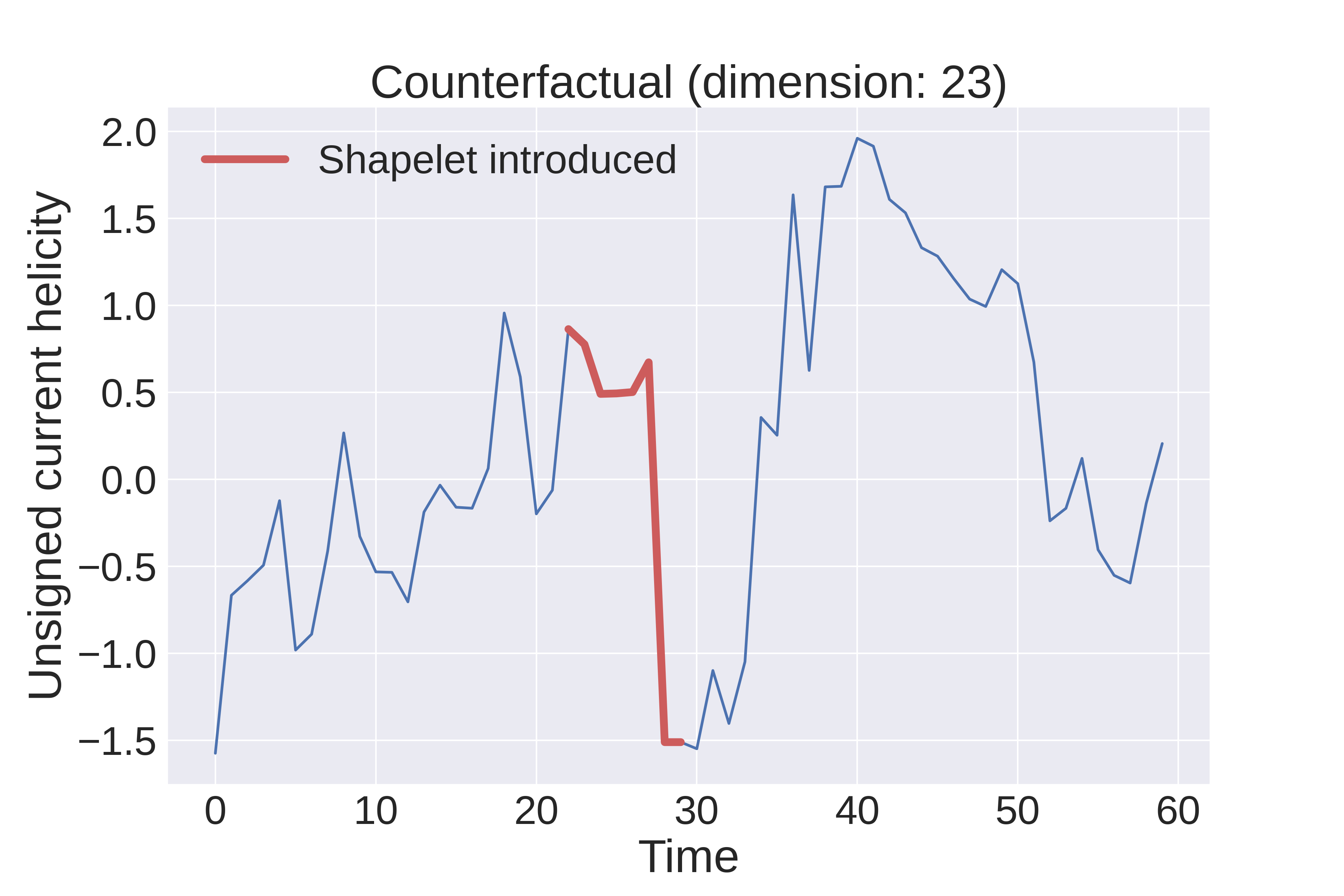}
        \caption{}
    \end{subfigure}
    \caption{Counterfactual generation using target class-shapelet introduction}
\end{figure}

In Figure 4, we show 4 more counterfactuals generated using SETS. For visualization purposes, we select counterfactuals that required perturbations at the level of one time series dimension only. However, SETS perturbed an average of 2.62 dimensions through the testing set. On the other hand, CoMTE substitutes entire dimensions to generate counterfactuals, which makes them less qualitatively interpretable. Moreover, in this experiment, significantly more dimensions were perturbed compared to SETS. The original univariate version of NG perturbs all time steps. Therefore, it suffers from the same problems as CoMTE in the multivariate adaptation.

\begin{figure}[t!]
    \centering
    \begin{subfigure}[t]{0.5\columnwidth}
        \centering
        \includegraphics[height=3.2cm]{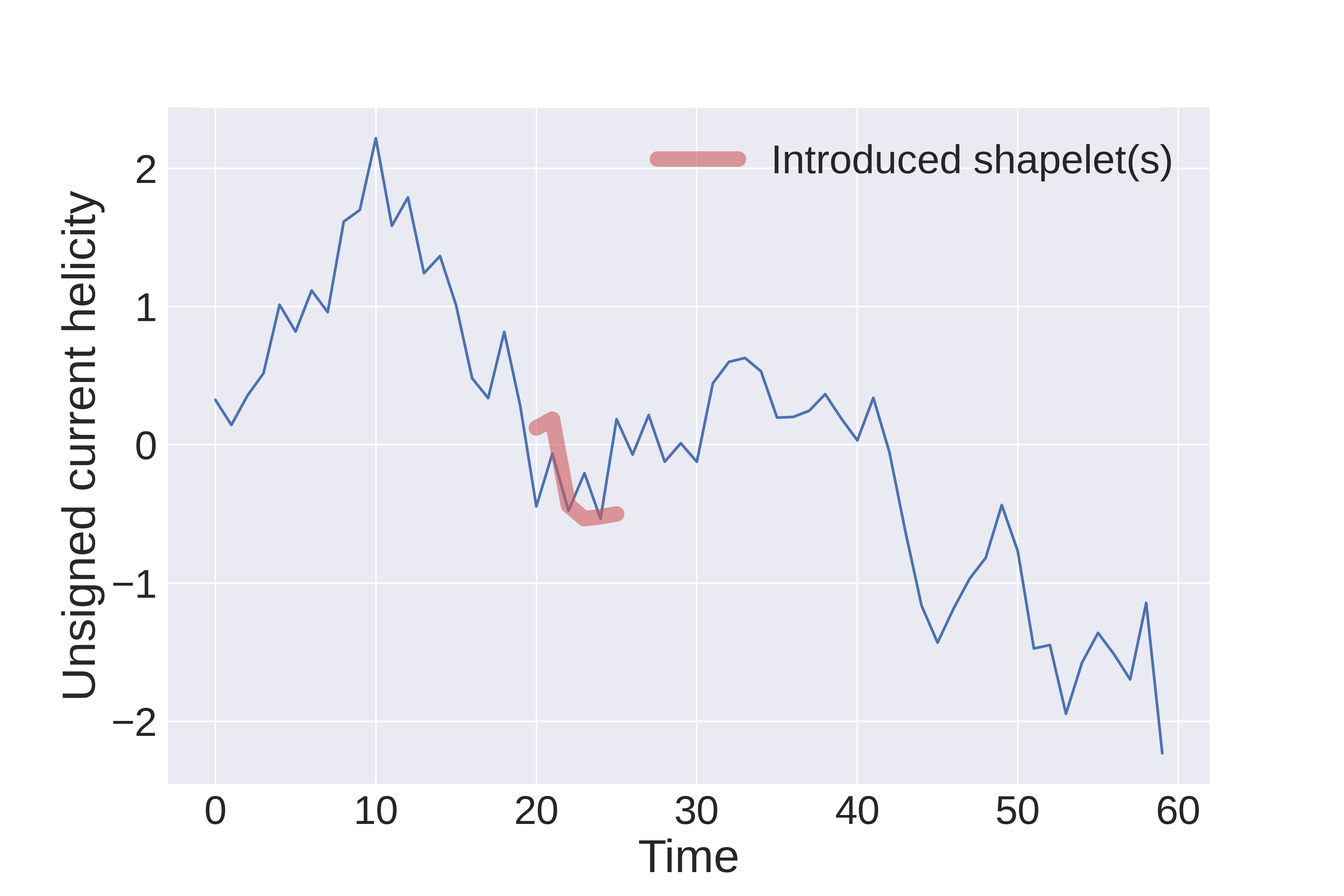}
        \caption{From class B/C to class Q}
    \end{subfigure}%
    ~ 
    \begin{subfigure}[t]{0.5\columnwidth}
        \centering
        \includegraphics[height=3.2cm]{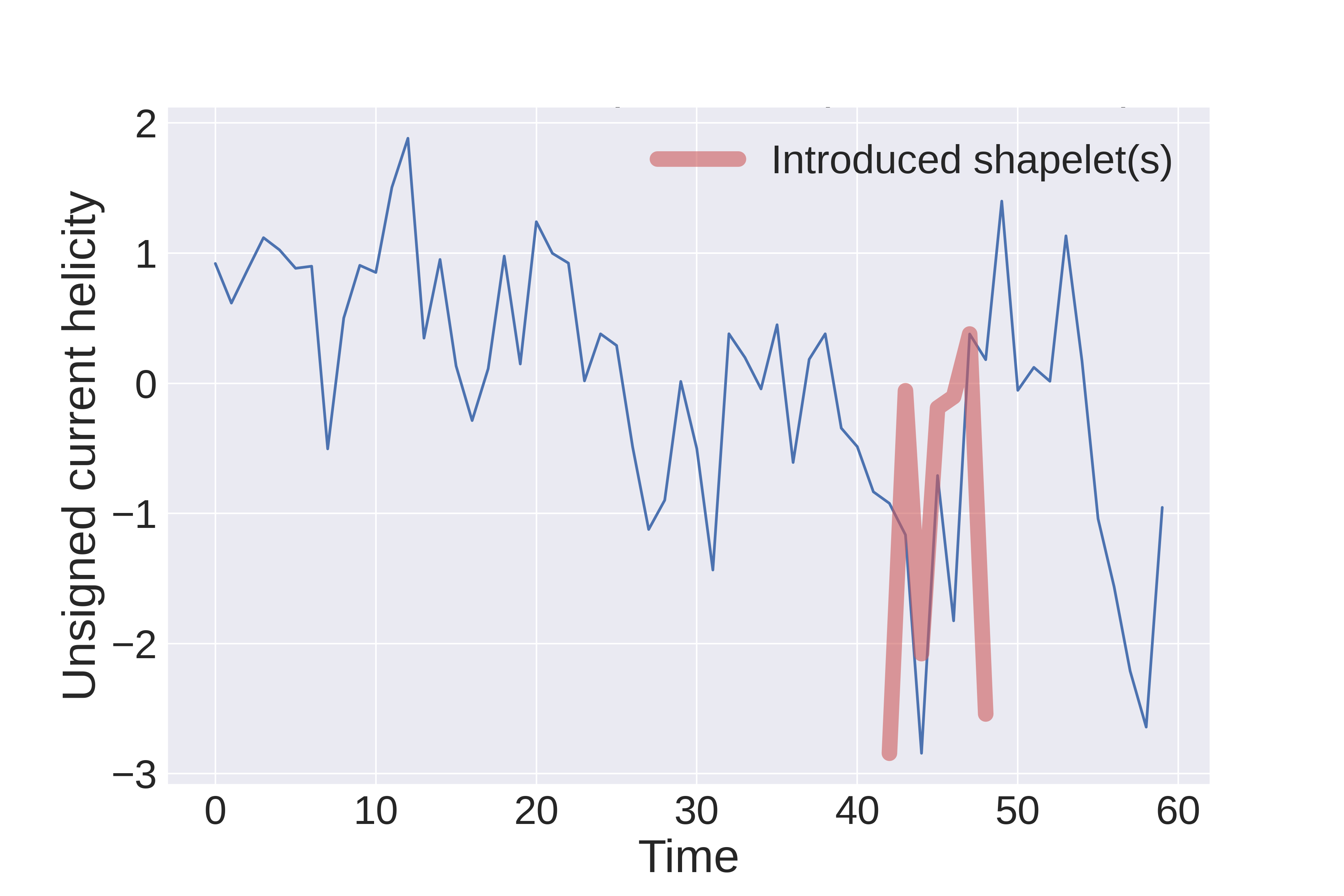}
        \caption{From class X to class B/C}
    \end{subfigure}%
    \\
    ~ 
    \begin{subfigure}[t]{0.5\columnwidth}
        \centering
        \includegraphics[height=3.2cm]{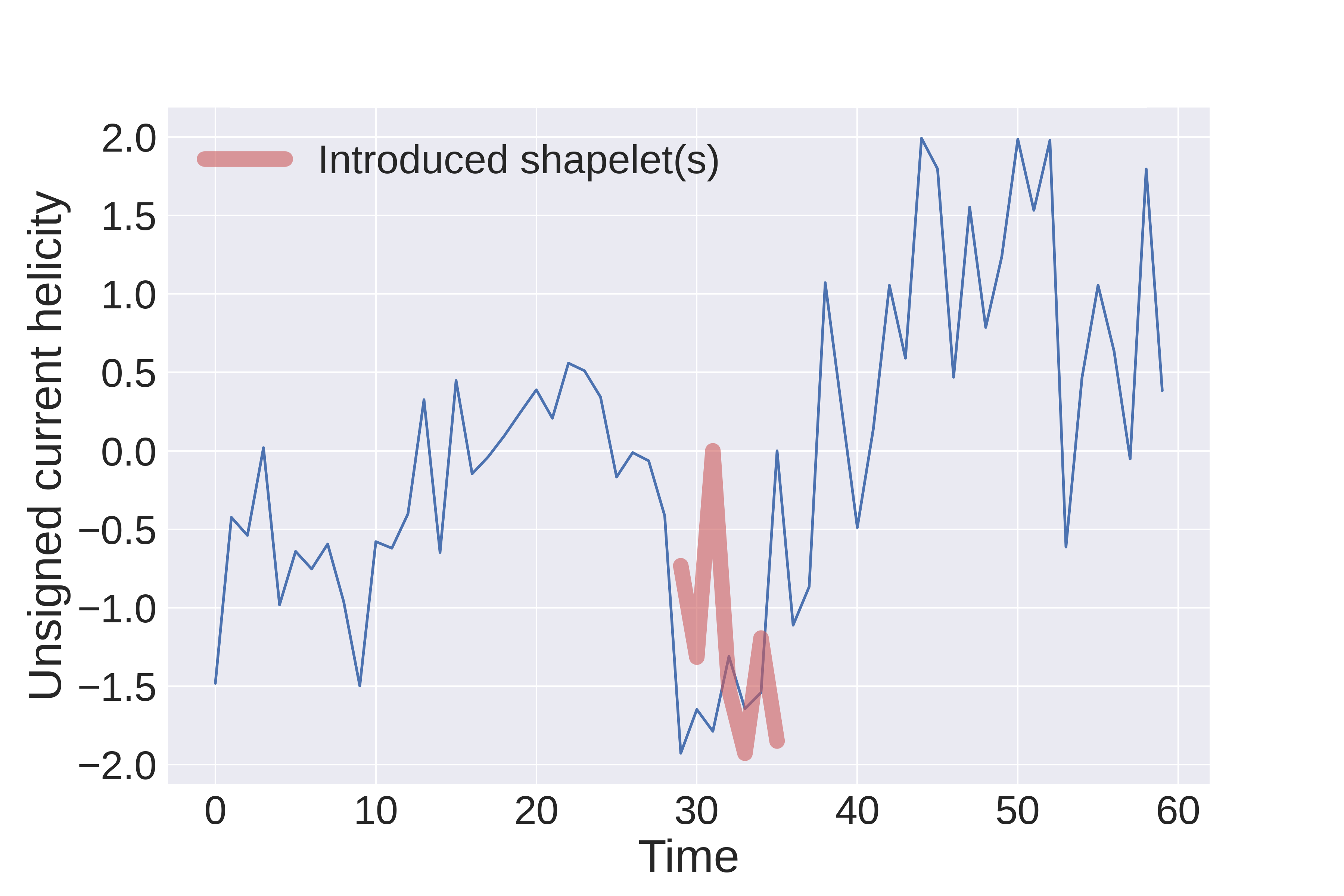}
        \caption{From class B/C to class M}
    \end{subfigure}%
    ~ 
    \begin{subfigure}[t]{0.5\columnwidth}
        \centering
        \includegraphics[height=3.2cm]{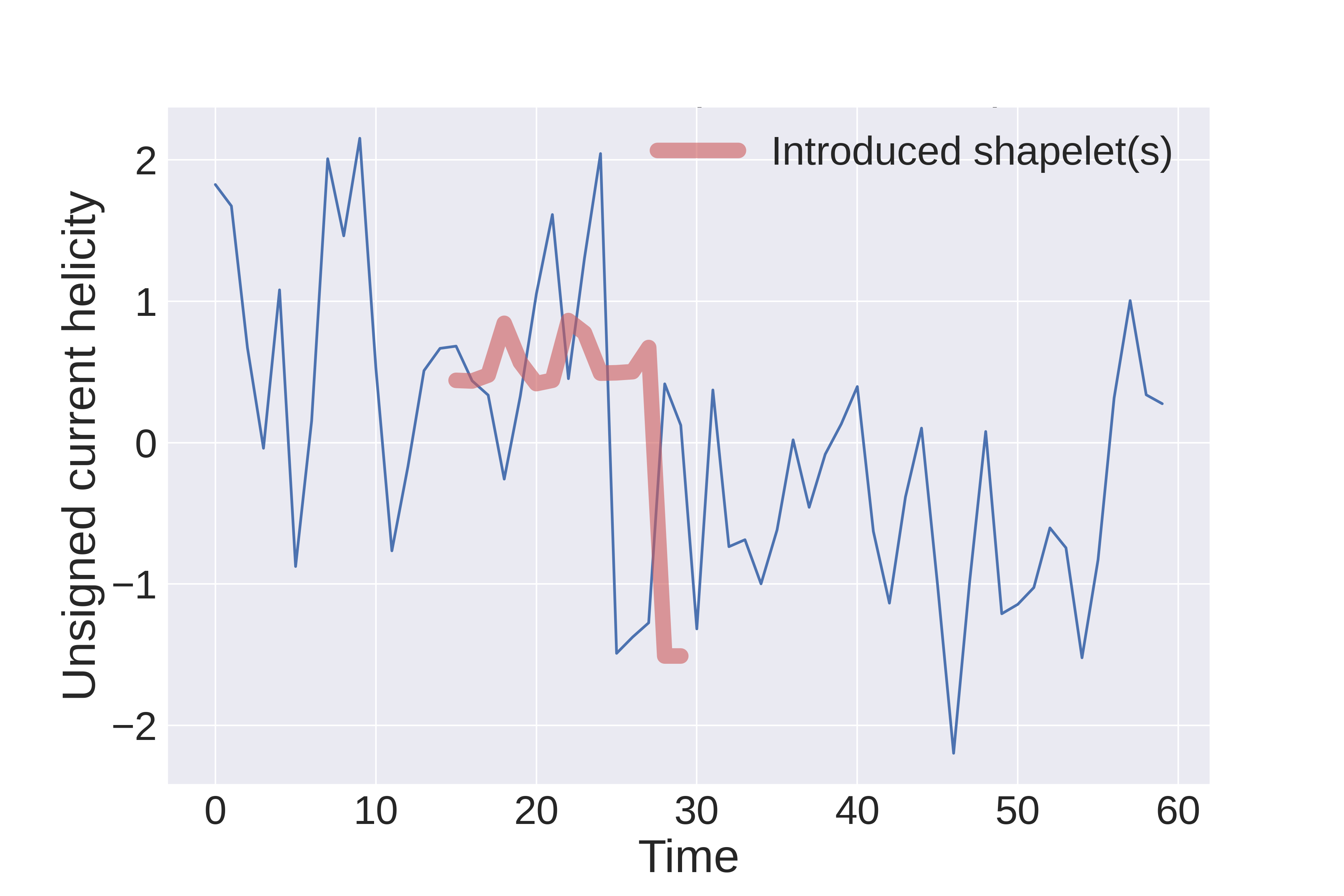}
        \caption{From class B/C to class X}
    \end{subfigure}
    \caption{Counterfactual explanations generated using SETS}
\end{figure}

\subsection{Quantitative Evaluation}
Out of the 1221 counterfactuals generated by NG, 11 are largely out-of-distribution. This is an expected outcome, as NG keeps incrementing the DBA weights until a counterfactual found. Therefore, the average evaluation measure values are highly affected by these 11 samples. In order, to fairly compare the three methods, we present both the means and medians of the the proximity and sparsity measures, and we show the plausibility results with and without these 11 outliers. However, it is important to note that 0.9\% of NG counterfactuals are significantly worse than average.

\subsubsection{Proximity}
As shown in Table 2, the counterfactuals generated by SETS are significantly closer to the original instances than those created using NG and CoMTE. This can be observed in all three metrics: the Manhattan distance ($L_{1}$-norm), the Euclidian distance ($L_{2}$-norm), and the $L_{\inf}$-norm. This means that not only are the counterfactuals closer to the original instances but that the perturbations are also much smaller in magnitudes; which suggests that NG and CoMTE counterfactuals contain more pronounced spikes. Table 2 also shows that NG produces closer counterfactuals than CoMTE, except for the 11 outliers discussed above (very high mean and lower median compared to CoMTE's).

\begin{table}[]
\centering
\caption{Proximity comparison}
\resizebox{\columnwidth}{!}{\begin{tabular}{c|cc|cc|cc|}
\cline{2-7}
                            & \multicolumn{2}{c|}{L1-norm}                          & \multicolumn{2}{c|}{L2-norm}                          & \multicolumn{2}{c|}{Linf-norm}                       \\ \cline{2-7} 
                            & \multicolumn{1}{c|}{Mean}            & Median         & \multicolumn{1}{c|}{Mean}            & Median         & \multicolumn{1}{c|}{Mean}            & Median        \\ \hline
\multicolumn{1}{|c|}{NG}    & \multicolumn{1}{c|}{1.42 x $10^{12}$} & 836.68         & \multicolumn{1}{c|}{1.38 x $10^{12}$} & 28.50          & \multicolumn{1}{c|}{1.37 x $10^{12}$} & 4.17          \\ \hline
\multicolumn{1}{|c|}{CoMTE} & \multicolumn{1}{c|}{2132.17}         & 2148.37        & \multicolumn{1}{c|}{60.81}           & 6.95           & \multicolumn{1}{c|}{6.96}            & 6.96          \\ \hline
\multicolumn{1}{|c|}{SETS}  & \multicolumn{1}{c|}{\textbf{90.64}}  & \textbf{86.41} & \multicolumn{1}{c|}{\textbf{10.11}}  & \textbf{10.95} & \multicolumn{1}{c|}{\textbf{2.64}}   & \textbf{2.47} \\ \hline
\end{tabular}}
\end{table}

\subsubsection{Sparsity}
We compute sparsity as the total number of perturbed time steps throughout all time series dimensions. Again, Table 3 shows that the counterfactual explanations generated by SETS are superior to NG's and CoMTE's. 

\begin{table}[]
\centering
\caption{Sparsity comparison}
\begin{tabular}{c|cc|}
\cline{2-3}
                            & \multicolumn{2}{c|}{Sparsity}                      \\ \cline{2-3} 
                            & \multicolumn{1}{c|}{Mean}          & Median        \\ \hline
\multicolumn{1}{|c|}{NG}    & \multicolumn{1}{c|}{135.69}        & 130.01        \\ \hline
\multicolumn{1}{|c|}{CoMTE} & \multicolumn{1}{c|}{159.38}        & 159.65        \\ \hline
\multicolumn{1}{|c|}{SETS}  & \multicolumn{1}{c|}{\textbf{9.29}} & \textbf{8.11} \\ \hline
\end{tabular} 
\end{table}

\subsubsection{Plausibility}
Table 4 shows that all plausibility measures consistently point to the fact that SETS generates significantly more plausible counterfactuals. The values in the table represent the percentage of out-of-distribution counterfactuals. The OC-SVM method on both the raw time space and the matrix profile favorises NG's counterfactuals over those generated by CoMTE. However, the IF and LOF methods have detected fewer outliers among CoMTE's.

\begin{table}[]
\centering
\caption{Plausibility comparison (percentage of out-of-distribution counterfactuals)}
\begin{tabular}{ll|l|l|l|}
\cline{3-5}
                                                 &                 & NG    & CoMTE        & SETS           \\ \hline
\multicolumn{1}{|l|}{\multirow{2}{*}{IF}}        & w/ NG outliers  & 0.28  & 0.004        & \textbf{0.0}   \\ \cline{2-5} 
\multicolumn{1}{|l|}{}                           & w/o NG outliers & 0.20  & 0.005        & \textbf{0.0}   \\ \hline
\multicolumn{1}{|l|}{\multirow{2}{*}{LOF}}       & w/ NG outliers  & 0.05  & \textbf{0.0} & \textbf{0.0}   \\ \cline{2-5} 
\multicolumn{1}{|l|}{}                           & w/o NG outliers & 0.04  & \textbf{0.0} & \textbf{0.0}   \\ \hline
\multicolumn{1}{|l|}{\multirow{2}{*}{OC-SVM}}    & w/ NG outliers  & 0.08  & 0.82         & \textbf{0.06}  \\ \cline{2-5} 
\multicolumn{1}{|l|}{}                           & w/o NG outliers & 0.06  & 0.82         & \textbf{0.06}  \\ \hline
\multicolumn{1}{|l|}{\multirow{2}{*}{OC-SVM MP}} & w/ NG outliers  & 0.076 & 0.816        & \textbf{0.057} \\ \cline{2-5} 
\multicolumn{1}{|l|}{}                           & w/o NG outliers & 0.065 & 0.816        & \textbf{0.058} \\ \hline
\end{tabular}
\end{table}

\section{Conclusion}
In this work, we proposed SETS, a model agnostic MTS counterfactual explanation algorithm. SETS makes use of shapelets to introduce meaningful perturbations to the original dataset instance, and to create highly interpretable counterfactuals. In particular, the perturbations introduced by SETS are contiguous, which makes the counterfactual explanations more plausible than the sparse ones generated by CoMTE. In addition, the use of shapelets provides SETS with the added quality of visual explainability. Indeed, plotting the counterfactuals and visualizing the perturbations can provide important insight as to which shapelets influenced the black-box model decision. We tested our approach on a real-life solar flare prediction dataset using a neural network as the black-box model and compared it to two state-of-the-art time series counterfactual generation algorithms, including the only one specifically developed for MTS. The results show that SETS' counterfactuals are superior in terms of proximity, sparsity, and plausibility, with the additional visual interpretability edge. In the future, we would like to experiment with an optimization-based approach, guided by a shapelet-based loss function.

\bibliography{./library}

\end{document}